\documentclass[conference]{IEEEtran}
\IEEEoverridecommandlockouts
\usepackage{cite}
\usepackage{amsmath,amssymb,amsfonts}
\usepackage{algorithmic}
\usepackage{graphicx}
\usepackage{textcomp}
\usepackage{xcolor}
\usepackage{booktabs}
\usepackage{caption} 
\usepackage{capt-of}
\usepackage{multirow}
\usepackage{threeparttable}
\usepackage{graphicx}
\usepackage{subcaption}
\usepackage{url}
\usepackage{acronym}
\usepackage{hyperref}
\hypersetup{
  colorlinks=true,
  linkcolor=blue,   
  citecolor=green,  
  urlcolor=blue  
}
\usepackage[linesnumbered,ruled,vlined]{algorithm2e} 
\usepackage{siunitx}
\def\BibTeX{{\rm B\kern-.05em{\sc i\kern-.025em b}\kern-.08em
    T\kern-.1667em\lower.7ex\hbox{E}\kern-.125emX}}
\begin{document}

\title{Parameter-Efficient Domain Adaptation of Physics-Informed Self-Attention based GNNs for AC Power Flow Prediction\\
}

\author{
    \IEEEauthorblockN{
        \textit{Redwanul Karim\textsuperscript{1}}\thanks{Corresponding author: \texttt{redwanul.karim@fau.de}. This work was conducted within the scope of the research project \textit{GridAssist} and was supported through the ``OptiNetD'' funding initiative by the German Federal Ministry for Economic Affairs and Energy (BMWE) as part of the 8\textsuperscript{th} Energy Research Programme. Source code and experimental configurations are available at \url{https://github.com/night-fury-me/efficient-graph-pf}.}, 
        \textit{Changhun Kim\textsuperscript{1}}, 
        \textit{Timon Conrad\textsuperscript{2}}, 
         \textit{Nora Gourmelon\textsuperscript{1}}, 
        \textit{Julian Oelhaf\textsuperscript{1}}, 
        \textit{David Riebesel\textsuperscript{2}}, \\
        \textit{Tomás Arias-Vergara\textsuperscript{1}}, 
        \textit{Andreas Maier\textsuperscript{1}}, 
        \textit{Johann Jäger\textsuperscript{2}},
        \textit{Siming Bayer\textsuperscript{1}}
    }
    \IEEEauthorblockA{
        \textsuperscript{1}Pattern Recognition Lab, Friedrich-Alexander-Universität Erlangen-Nürnberg, Erlangen, Germany\\
        \textsuperscript{2}Institute of Electrical Energy Systems, Friedrich-Alexander-Universität Erlangen-Nürnberg, Germany \\
    }
}

\maketitle

\acrodef{GNN}{Graph Neural Network}
\acrodef{AC-PF}{AC Power Flow}
\acrodef{NR}{Newton--Raphson}
\acrodef{LoRA}{Low-Rank Adaptation}
\acrodef{OPF}{Optimal Power Flow}

\begin{abstract}
\noindent Accurate \ac{AC-PF} prediction under domain shift is critical when models trained on medium-voltage (MV) grids are deployed on high-voltage (HV) networks. Existing physics-informed graph neural solvers typically rely on full fine-tuning for cross-regime transfer, incurring high retraining cost and offering limited control over the stability--plasticity trade-off between target-domain adaptation and source-domain retention. We study parameter-efficient domain adaptation for physics-informed self-attention based \ac{GNN}, encouraging Kirchhoff-consistent behavior via a physics-based loss while restricting adaptation to low-rank updates. Specifically, we apply \ac{LoRA} to attention projections with selective unfreezing of the prediction head to regulate adaptation capacity. This design yields a controllable efficiency--accuracy trade-off for physics-constrained inverse estimation under voltage-regime shift. Across multiple grid topologies, the proposed \texttt{LoRA+PHead} adaptation recovers near--full fine-tuning accuracy with a target-domain RMSE gap of $\boldsymbol{2.6\times10^{-4}}$ while reducing the number of trainable parameters by $\boldsymbol{85.46\%}$. The physics-based residual remains comparable to full fine-tuning; however, relative to Full FT, \texttt{LoRA+PHead} reduces MV source retention by 4.7 percentage points (17.9\% vs.\ 22.6\%) under domain shift, while still enabling parameter-efficient and physically consistent \ac{AC-PF} estimation.
\end{abstract}


\section{Introduction}
\label{sec:intro}

Accurate and fast solution of the AC power flow (AC-PF) equations underpins state estimation, contingency screening, and real-time security assessment in modern power systems \cite{stott1974fast, elfergany2025reviews, skolfield2022operations}. Increasing renewable penetration, volatile operating points, and frequent topology changes primarily stress classical Newton--Raphson and fast-decoupled solvers in terms of computational throughput: while per-instance AC-PF solves remain reliable for updated $\mathbf{Y}_{\text{bus}}$ and injections, repeated sparse Jacobian factorizations become a bottleneck in time-critical large-scale screening pipelines \cite{tinney1967direct, hailu2023techniques}. Learning-based surrogates offer a complementary path by incurring a one-time training cost and enabling fast per-scenario inference across many operating conditions.

\ac{GNN} model the power grid as a graph and learn via neighborhood message passing along electrical connectivity, inducing a topology-aware and permutation-equivariant inductive bias that enables generalization across network sizes and topological configurations \cite{diehl2019warm, ringsquandl2021deep, deihim2024initial}. Message-passing \ac{GNN} solvers and warm-start models---where neural predictors initialize iterative AC power flow or \ac{OPF} solvers to accelerate convergence---achieve strong performance for \ac{AC-PF} and \ac{OPF} \cite{donon2020neural, lin2024powerflownet, lopez2023power}, while \ac{GNN} improve expressivity via attention-based aggregation \cite{velickovic2018graph}. However, purely data-driven \ac{GNN} may violate Kirchhoff's laws under distribution shift \cite{hu2021physics}. Physics-informed \ac{GNN} embed the governing equations into the learning objective, improving extrapolation and physical feasibility \cite{lopez2023power, kim2025physics}, but are typically trained and deployed in full precision, entailing higher computational and memory requirements.

From a deployment perspective, inference latency, memory footprint, and retraining cost are first-order constraints. Although quantization can reduce arithmetic cost and memory bandwidth \cite{banner2018scalable}, it is difficult to apply reliably to \ac{GNN} due to degree-dependent aggregation and irregular dataflow \cite{tailor2020degree}. We therefore focus on parameter-efficient fine-tuning as a practical means to reduce adaptation cost at deployment. At the same time, domain shift across grid regimes poses a central challenge: despite the topology-aware inductive bias of \ac{GNN}, models trained on medium-voltage (MV) networks degrade when transferred to high-voltage (HV) systems due to changes in topology and operating statistics \cite{ringsquandl2021deep, hamann2024foundation}. Existing domain adaptation strategies typically rely on full fine-tuning, which is computationally expensive and induces substantial source-domain forgetting \cite{wu2020domain, sun2016deep}, motivating approaches that explicitly control the stability--plasticity trade-off. Augmenting edge features with detailed line parameters (e.g., $R/X$) to better capture cross-regime shifts would further increase model size, strengthening the case for parameter-efficient adaptation.

\noindent\textbf{Relation to Prior \ac{GNN}-Based Power-Flow Solvers.}
Prior work establishes \ac{GNN} as effective AC-PF surrogates \cite{donon2020neural, ringsquandl2021deep} and shows that physics-informed training improves physical consistency \cite{hu2021physics, dejongh2022physics, jeddi2021physics, kim2025physics}. However, cross-regime adaptation from MV to HV remains underexplored. This shift entails changes in voltage levels (partially normalized in p.u.), larger and different topologies, modified admittance matrices $\mathbf{Y}_{\text{bus}}$, and different line $R/X$ ratios, which alter power-flow characteristics and reactive power demands. While \ac{GNN}s encode connectivity, they do not inherently account for such parameter shifts. We study parameter-efficient adaptation of physics-informed \acp{GNN} for AC power flow under MV$\to$HV regime shift. To the best of our knowledge, this is the first \emph{systematic} analysis of low-rank adaptation in this setting, evaluating accuracy, physics-residual feasibility, optimization stability, and source-domain retention beyond the full fine-tuning baseline.

\noindent\textbf{Contributions.}
(i) We introduce low-rank adaptation of attention projections with structurally motivated, selective unfreezing of the per-step iterative-refinement heads for parameter-efficient transfer in physics-informed GNNs;
(ii) we achieve near-full fine-tuning accuracy with $>85\%$ fewer trainable parameters while maintaining comparable physics residuals;
(iii) we characterize the stability--plasticity trade-off via source-retention and few-shot scaling, and establish the efficiency--accuracy Pareto frontier for cross-regime transfer under limited supervision.

\section{Methodology}

\subsection{Problem Formulation}

We model a power grid as a graph $\mathcal{G}=(\mathcal{V},\mathcal{E})$, where nodes $i\in\mathcal{V}$ represent buses (PQ, PV, and slack) and edges $(i,j)\in\mathcal{E}$ represent transmission lines. Each node is associated with input features $\mathbf{x}_i\in\mathbb{R}^{d_x}$ encoding net active/reactive power injections, bus type indicators (PQ/PV/slack), and voltage setpoints for PV/slack; each edge with features $\mathbf{e}_{ij}\in\mathbb{R}^{d_e}$ encoding line parameters (e.g., $R/X$, susceptance). Let $\mathbf{X}=\{\mathbf{x}_i\}_{i\in\mathcal{V}}$ and $\mathbf{E}=\{\mathbf{e}_{ij}\}_{(i,j)\in\mathcal{E}}$ denote the collections of node and edge features.

The learning objective is to approximate the AC power-flow operator by a parametric predictor
\begin{equation}
f_\theta:\; (\mathcal{G}, \mathbf{X}, \mathbf{E}) \;\mapsto\; \hat{\mathbf{Y}},
\qquad 
\hat{\mathbf{Y}} = \{ (\hat{V}_i, \hat{\theta}_i) \}_{i\in\mathcal{V}},
\end{equation}
where $(\hat{V}_i, \hat{\theta}_i)$ denote predicted voltage magnitudes and phase angles for all buses, with the slack complex voltage is fixed to their setpoints when computing loss and evaulation metrics.

Given training samples $(\mathcal{G}^{(k)}, \mathbf{X}^{(k)}, \mathbf{E}^{(k)}, \mathbf{Y}^{(k)}) \sim \mathcal{D}_s$, we train $f_\theta$ on the source domain by minimizing the combined data-fitting and physics-informed objective:
\begin{equation}
\theta_s = \arg\min_{\theta \in \Theta}\; 
\mathbb{E}_{(\mathcal{G},\mathbf{X},\mathbf{E},\mathbf{Y}) \sim \mathcal{D}_s}
\!\left[ 
\mathcal{L}_{\text{data}} + \lambda_{\mathrm{PF}}\,\mathcal{L}_{\mathrm{PF}}
\right],
\end{equation}
where the data-fitting loss $\mathcal{L}_{\text{data}}$ is defined by the RMSE in Eq.~\eqref{eq:rmse} and the physics-informed residual loss $\mathcal{L}_{\mathrm{PF}}$ is given by Eq.~\eqref{eq:lpf}. 
Here, $\lambda_{\mathrm{PF}} \geq 0$ is a scalar weighting coefficient that controls the trade-off between empirical data fidelity and adherence to Kirchhoff-consistent power-flow constraints, with larger values encourages stronger physical consistency at the potential cost of reduced data fit.

We consider a domain-adaptation setting with source and target distributions $\mathcal{D}_s \neq \mathcal{D}_t$, corresponding to different voltage regimes and operating statistics. Let $\theta_s$ denote the source-trained parameters. The adapted parameters are obtained by restricting updates to a parameter-efficient adaptation subspace $\mathcal{A}(\theta_s)$:
\begin{equation}
\begin{aligned}
\theta_t
&= \arg\min_{\theta \in \mathcal{A}(\theta_s)}
\mathbb{E}_{(\mathcal{G},\mathbf{X},\mathbf{E},\mathbf{Y}) \sim \mathcal{D}_t}
\Big[
\mathcal{L}_{\text{data}} + \lambda_{\mathrm{PF}}\,\mathcal{L}_{\mathrm{PF}}
\Big],
\end{aligned}
\end{equation}
where $\mathcal{A}(\theta_s)$ denotes the parameter-efficient adaptation subspace (Sec.~\ref{sec:peft}), and $\theta_s, \theta_t$ denote the source and adapted model parameters, respectively.

\subsection{Edge-Aware Self-Attention GNN}

We employ a Transformer-style multi-head self-attention \ac{GNN} with explicit edge-conditioned \emph{attention biases}, following~\cite{kim2025physics}. Let $\mathbf{h}_i^{(\ell)}\in\mathbb{R}^{d\times 1}$ denote the column-vector embedding of node $i$ at layer $\ell$, and let $\mathcal{N}(i)$ be its one-hop neighbors. For head $m=1,\dots,M$ with per-head dimensionality $d_h$, we compute
\begin{equation}
\mathbf{q}_i^{m}=\mathbf{W}_Q^{m}\mathbf{h}_i^{(\ell)},\quad
\mathbf{k}_j^{m}=\mathbf{W}_K^{m}\mathbf{h}_j^{(\ell)},\quad
\mathbf{v}_j^{m}=\mathbf{W}_V^{m}\mathbf{h}_j^{(\ell)},
\end{equation}
where $\mathbf{W}_Q^{m}, \mathbf{W}_K^{m}, \mathbf{W}_V^{m}\in\mathbb{R}^{d_h\times d}$. Line features $\boldsymbol{\ell}_{ij}\in\mathbb{R}^{d_e\times 1}$ (e.g., admittance parameters) are mapped to a scalar, head-specific edge bias
\begin{equation}
\beta_{ij}^{m} = f_{\text{edge}}^{m}(\boldsymbol{\ell}_{ij}), \qquad 
f_{\text{edge}}^{m}:\mathbb{R}^{d_e\times 1}\rightarrow\mathbb{R},
\end{equation}
which augments the scaled dot-product attention logit:
\begin{equation}
s_{ij}^{m}
= \frac{(\mathbf{q}_i^{m})^\top \mathbf{k}_j^{m}}{\sqrt{d_h}} + \beta_{ij}^{m}, 
\qquad
\alpha_{ij}^{m}=\frac{\exp(s_{ij}^{m})}{\sum_{k\in\mathcal{N}(i)}\exp(s_{ik}^{m})}.
\end{equation}

\noindent The head-wise aggregation is
\begin{equation}
\mathbf{h}_i^{m}=\sum_{j\in\mathcal{N}(i)}\alpha_{ij}^{m}\mathbf{v}_j^{m}, 
\qquad \mathbf{h}_i^{m}\in\mathbb{R}^{d_h\times 1},
\end{equation}
and concatenation over heads followed by a linear projection yields $\mathbf{h}_i^{(\ell+1)}\in\mathbb{R}^{d\times 1}$. Injecting admittance-derived features as edge-dependent biases yields a directional, state-dependent propagation operator (generally $\alpha_{ij}\neq\alpha_{ji}$) that better captures anisotropic electrical couplings and improves robustness under voltage-regime shifts and topology changes.

\subsection{Parameter-Efficient Domain Adaptation}
\label{sec:peft}

To enable controlled \emph{model adaptation capacity} under domain shift, we restrict target-domain updates to low-rank perturbations of selected attention projection matrices, as summarized in Alg.~\ref{alg:adaptation}. Specifically, we apply low-rank adaptation to the query, key, and value projections ($\mathbf{W}_Q, \mathbf{W}_K, \mathbf{W}_V$). For any adapted projection $\mathbf{W}\in\mathbb{R}^{d_{\text{out}}\times d_{\text{in}}}$, we reparameterize
\begin{equation}
\mathbf{W}'=\mathbf{W}+\Delta\mathbf{W}, 
\qquad 
\Delta\mathbf{W}=\frac{\alpha}{r}\,\mathbf{A}\mathbf{B},
\end{equation}
with $\mathbf{A}\in\mathbb{R}^{d_{\text{out}}\times r}$, $\mathbf{B}\in\mathbb{R}^{r\times d_{\text{in}}}$, and $r \ll \min(d_{\text{in}},d_{\text{out}})$, following \cite{hu2022lora}. We use $r=r_{\text{lora}}$ and $\alpha=\alpha_{\text{lora}}$; all base parameters are frozen and only $\{\mathbf{A},\mathbf{B}\}$ are optimized. The prediction module comprises three per-step heads ($\theta$-, $V$-, message-) replicated over $K$ iterative refinement steps implementing a line-search correction operator over the AC-PF residual \cite{kim2025physics}; these are zero-initialized, so regime-specific output rescaling (per-unit angle/voltage range under MV$\to$HV) is structurally absorbed by head weights, which low-rank attention deltas cannot reproduce. Selective head unfreezing is therefore structurally necessary, not merely additive, and jointly with LoRA controls the stability--plasticity trade-off. The strategy is backbone-agnostic: it requires only named Q/K/V projections and a separable head, and extends directly to transformer regressors and edge-conditioned MPNNs with attention readout.

\begin{algorithm}[t]
\SetKwInOut{Input}{Input}
\SetKwInOut{Output}{Output}
\SetKwInOut{Initialize}{Initialize}
\Input{Source dataset $\mathcal{D}_s$; Target dataset $\mathcal{D}_t$; GNN model $f_{\boldsymbol{\phi}}$; Physics weights $\lambda_P, \lambda_Q$; LoRA rank $r$}
\Output{Adapted model $\hat{f}_{\boldsymbol{\phi}}$}
\Initialize{Initialize base parameters $\boldsymbol{\phi}$; Initialize LoRA matrices $\mathbf{A}_\ell \sim \mathcal{N}(0,\sigma^2)$, $\mathbf{B}_\ell \leftarrow \mathbf{0}$}
\ForEach{batch from $\mathcal{D}_s$}{
  Predict $(\hat{\mathbf{V}}, \hat{\boldsymbol{\theta}})$\;
  Compute composite loss $\mathcal{L}$\;
  Update $\boldsymbol{\phi}$ via gradient descent\;
}
Freeze base parameters $\boldsymbol{\phi}$\;
\ForEach{batch from $\mathcal{D}_t$}{
  Apply low-rank updates $\mathbf{W}'_\ell \leftarrow \mathbf{W}_\ell + \mathbf{A}_\ell \mathbf{B}_\ell$\;
  Predict outputs and compute $\mathcal{L}$\;
  Update only $\{\mathbf{A}_\ell, \mathbf{B}_\ell\}$ and the prediction head\;
}
\Return $\hat{f}_{\boldsymbol{\phi}}$
\caption{Parameter-Efficient Domain Adaptation of Physics-Informed Self-Attention based \ac{GNN}}
\label{alg:adaptation}
\end{algorithm}

\subsection{Computational Complexity}

Let $N=|\mathcal{V}|$, $E=|\mathcal{E}|$, $d$ the hidden dimension, and $H$ the number of attention heads. The edge-aware multi-head self-attention layer has dominant cost $\mathcal{O}(N d^2 + E d)$ ($H$ absorbed for fixed $d$). LoRA adds $\mathcal{O}(N r d)$ with $r\ll d$, which is lower order and leaves asymptotic complexity unchanged. Parameter-wise, full fine-tuning updates $\mathcal{O}(d^2)$ per projection while LoRA reduces this to $\mathcal{O}(r(d_{\text{in}}+d_{\text{out}}))$, explaining the observed efficiency--accuracy trade-offs.

\begin{table*}[t]
\centering

\begin{tabular}{@{}p{0.68\textwidth}@{\hspace{5pt}}p{0.28\textwidth}@{}}

\begin{minipage}[t]{\linewidth}
\centering
\begin{threeparttable}
  \caption{Cross-regime AC-PF prediction performance (MV$\rightarrow$HV).}
  \label{tab:main_results}

  \setlength{\tabcolsep}{2.3pt}
  \renewcommand{\arraystretch}{1.15}

  \begin{tabular}{lcccccc}
    \toprule
    \multirow{2}{*}{\textbf{Method}} 
    & \multicolumn{4}{c}{\textbf{Prediction Accuracy}} 
    & \multicolumn{2}{c}{\textbf{Efficiency}} \\
    \cmidrule(lr){2-5} \cmidrule(lr){6-7}
    & RMSE$_{\text{all}}$ $\downarrow$ 
    & RMSE$_V$ $\downarrow$ 
    & RMSE$_\theta$ ($^\circ$) $\downarrow$ 
    & $\mathcal{L}_{\text{PF}}$ $\downarrow$
    & $\mathcal{P}_{\text{reduced}}$ (\%) $\uparrow$ 
    & $\mathcal{R}_{\text{ret}}$ (\%) $\uparrow$ \\
    \midrule
    ZeroShot (Base)     
    & $1.65 \times 10^{-2}$ 
    & $1.71 \times 10^{-3}$ 
    & $9.38 \times 10^{-1}$ 
    & $2.57$ 
    & \textemdash{} 
    & \textemdash{} \\
    
    \textbf{Full FT}       
    & $9.35 \times 10^{-4}$ 
    & $3.61 \times 10^{-4}$ 
    & $4.95 \times 10^{-2}$ 
    & $1.11$ 
    & $0\%$ 
    & $22.6\%$ \\
    \addlinespace
    \texttt{Head-only}
    & $1.38 \times 10^{-3}\,^{\ddagger}$
    & $4.19 \times 10^{-4}\,^{\ddagger}$
    & $7.51 \times 10^{-2}\,^{\ddagger}$
    & $\mathbf{1.11}\,^{\ddagger}$
    & $88.50\%$
    & $\mathbf{22.3}\%$ \\

    \texttt{LoRA-only}
    & $3.61 \times 10^{-3}$
    & $1.05 \times 10^{-3}$
    & $1.98 \times 10^{-1}$
    & $6.08$
    & $\mathbf{96.56}\%$
    & $17.3\%$ \\

    \texttt{LoRA+PHead}
    & $\mathbf{1.20 \times 10^{-3}}\,^{\ddagger}$
    & $\mathbf{3.53 \times 10^{-4}}\,^{\ddagger}$
    & $\mathbf{6.56 \times 10^{-2}}\,^{\ddagger}$
    & $1.21\,^{\ddagger}$
    & $85.46\%$
    & $17.9\%$ \\
    \bottomrule
  \end{tabular}

  \begin{tablenotes}[para, flushleft]
    \footnotesize
    Best results are in \textbf{bold}. $\downarrow$ $\uparrow$ indicates lower/higher is better.
    RMSE: Root Mean Square Error.
    $\mathcal{L}_{\text{PF}}$: Physics-informed PF loss.
    $\mathcal{P}_{\text{reduced}}$: Parameter reduction vs.\ Full FT.
    $\mathcal{R}_{\text{ret}}$: MV retention, defined as $\mathcal{R}_{\text{ret}} = 100 \cdot \mathrm{RMSE}_{\text{MV}}^{\text{meth}} / \mathrm{RMSE}_{\text{MV}}^{\text{base}}$.
    $^{\ddagger}p<0.01$ (paired Wilcoxon vs.\ Full FT over $N_{\text{wilcox}}=500$).
  \end{tablenotes}

\end{threeparttable}
\end{minipage}

&
\begin{minipage}[t]{\linewidth}
\centering
\centering
\captionsetup{font=footnotesize}


\captionof{table}{MV/HV parameter ranges.}
\label{tab:distribution}

\captionsetup{skip=2pt}

\scriptsize
\setlength{\tabcolsep}{2pt}
\renewcommand{\arraystretch}{1.05}

\begin{tabular}{@{}lcc@{}}
\toprule
\textbf{Parameter (unit)} & \textbf{MV} & \textbf{HV} \\
\midrule
Base voltage (kV)                & 10            & 110 \\
Base power (MVA)                 & 10            & 100 \\
Line length (km)                 & 1--20         & 1--50 \\
Series resistance ($\Omega$/km)  & 0.5--0.6       & 0.15--0.20 \\
Series reactance ($\Omega$/km)   & 0.30--0.35     & 0.35--0.45 \\
Shunt capacitance (nF/km)        & 8--14          & 8--10 \\
Active power (MW)                & $[-5,\,5]$     & $[-300,\,300]$ \\
Reactive power (MVAr)            & $[-2,\,2]$     & $[-150,\,150]$ \\
\bottomrule
\end{tabular}
  \begin{tablenotes}[para, flushleft]
    \footnotesize 
    MV/HV experimental setup: Each graph has \\ 4--32 buses; datasets contain 90{,}030 (MV) and \\45{,}030 (HV) samples; Train/Val/Test split 1:1:1
  \end{tablenotes}
\end{minipage}

\end{tabular}

\end{table*}

\begin{figure*}[t]
  \centering
  \newcommand{\subfigH}{0.14\textheight}

  \begin{subfigure}[t]{0.35\textwidth}
    \centering
    \includegraphics[width=\linewidth,height=\subfigH,keepaspectratio]{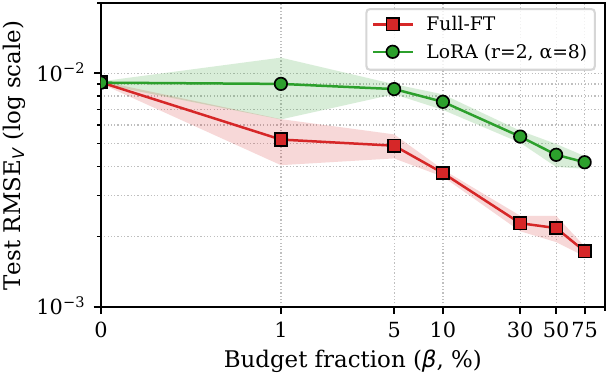}
    \caption{Few-shot target-domain adaptation vs. labeled fraction $\beta$.}
    \label{fig:fewshot_band}
  \end{subfigure} \hspace{0.5em}
  \begin{subfigure}[t]{0.40\textwidth}
    \centering
    \includegraphics[width=\linewidth,height=\subfigH,keepaspectratio]{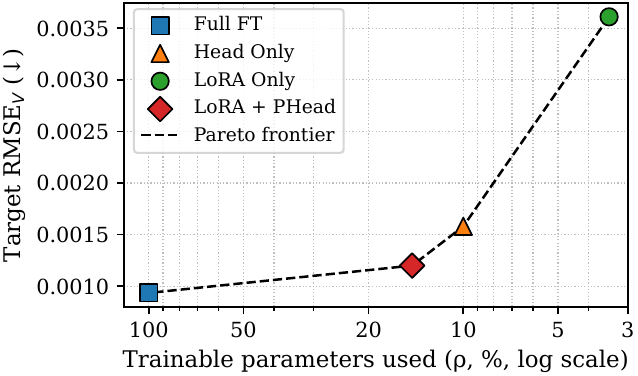}
    \caption{Efficiency--accuracy Pareto trade-off (RMSE vs. trainable parameters).}
    \label{fig:pareto_tradeoff}
  \end{subfigure}

  \vspace{0.5em}
  \newcommand{\subfigHH}{0.16\textheight}
  \begin{subfigure}[t]{0.35\textwidth}
    \centering
    \includegraphics[width=\linewidth,height=\subfigHH,keepaspectratio]{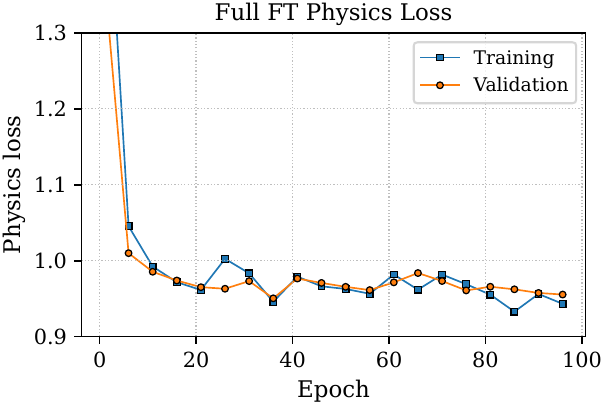}
    \caption{Physics-loss $\mathcal{L}_{\mathrm{PF}}$ during full fine-tuning.}
    \label{fig:physics_loss_full_ft}
  \end{subfigure} \hspace{0.5em}
  \begin{subfigure}[t]{0.40\textwidth}
    \centering
    \includegraphics[width=\linewidth,height=\subfigHH,keepaspectratio]{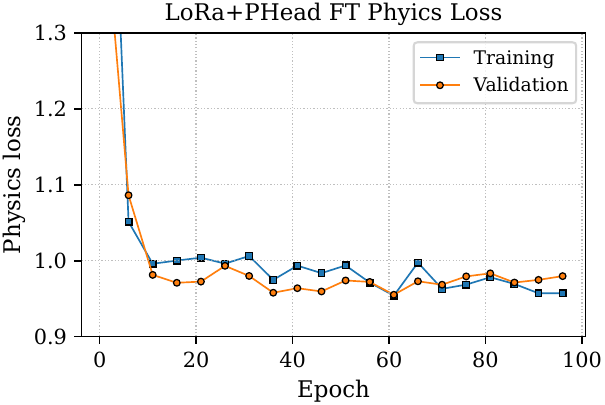}
    \caption{Physics-loss $\mathcal{L}_{\mathrm{PF}}$ during LoRA+PHead adaptation.}
    \label{fig:physics_loss_lora_head}
  \end{subfigure}

  \caption{Cross-regime adaptation (MV$\to$HV): few-shot scaling, efficiency--accuracy trade-off, and physics-loss dynamics for full fine-tuning and parameter-efficient adaptation.}
  \label{fig:main_composite}
\end{figure*}

\section{Dataset and Experimental Setup}
We synthesize physically plausible MV/HV datasets following Europe/Germany-typical overhead-line parameter ranges for series impedance and shunt charging (Table~\ref{tab:distribution}) \cite{oswald2005eev,kremens1996,theil2012}. Each sample is a connected graph $\mathcal{G}=(\mathcal{V},\mathcal{E})$ with one slack bus and remaining PV/PQ buses; lines are modeled with $\pi$-equivalent circuits. MV grids emphasize higher $R/X$ and shorter spans, whereas HV grids are reactance-dominated with longer spans.

Node features $\mathbf{x}_i$ encode net injections $(P_i^{\mathrm{set}}, Q_i^{\mathrm{set}})$, bus-type indicators, and voltage setpoints (for slack/PV buses); edge features $\mathbf{e}_{ij}$ encode line physics via per-edge biases using admittance features. Initial states use $|V_i^{(0)}|\!\in\![0.9,1.1]$ p.u. for slack/PV and $|V_i^{(0)}|\!=\!1.0$ p.u. for PQ buses with zero angle. Reference targets $\mathbf{Y}=\{(V_i^\ast,\theta_i^\ast)\}_{i\in\mathcal{V}}$ are obtained by Newton--Raphson on the bus-admittance matrix $Y_{\text{bus}}$ and setpoints $\mathbf{S}^{\mathrm{set}}=\{(P_i^{\mathrm{set}},Q_i^{\mathrm{set}})\}$ under slack/PV/PQ constraints; non-convergent or disconnected cases are discarded.

All quantities are expressed in per-unit using $(S_{\text{base}}, V_{\text{base}})$. Data are split into train/validation/test with a fixed seed. We train a physics-informed GNN with edge-aware self-attention ($d=8$, $d_{\text{hi}}=32$, $K=40$, 8 layers) using the combined RMSE loss (Eq.~\eqref{eq:rmse}) and physics residual loss (Eq.~\eqref{eq:lpf}) for 100 epochs (batch size 512) with AdamW (lr $10^{-4}$, weight decay $10^{-3}$) and cosine annealing with warm restarts. We set $r_{\text{lora}} = 2$ and $\alpha_{\text{lora}} = 8$ unless stated otherwise. GPU acceleration is used when available.

\section{Results and Analysis}
\label{sec:results}

We evaluate MV$\to$HV cross-regime adaptation under identical backbones and training budgets to test the central contribution that \emph{parameter-efficient low-rank adaptation recovers near--full fine-tuning accuracy under domain shift while preserving physical consistency and enabling controllable stability--plasticity trade-offs}. Prediction accuracy is measured by
\begin{equation}
\label{eq:rmse}
\mathcal{L}_{\text{data}}(\hat{\mathbf{y}}, \mathbf{y})
\;\triangleq\;
\mathrm{RMSE}(\hat{\mathbf{y}}, \mathbf{y})
\;=\;
\sqrt{\frac{1}{N}\sum_{i=1}^{N}\lVert \hat{\mathbf{y}}_i-\mathbf{y}_i \rVert_2^2},
\end{equation}
reported for voltage magnitude and phase angle. Physical feasibility is quantified by the AC power-flow residual loss

\begin{equation}
\label{eq:lpf}
\mathcal{L}_{\mathrm{PF}} = 
\sum_{t=0}^{T-1} \gamma^{T-1-t} 
\left(
\|\Delta P^{(t)}\|_2^2 + \|\Delta Q^{(t)}\|_2^2
\right),
\end{equation}

with $\Delta P^{(t)} = P^{\text{set}}-\Re\{V^t \odot (YV^t)^{*}\}$ and $\Delta Q^{(t)} = Q^{\text{set}}-\Im\{V^t \odot (YV^t)^{*}\}$, where $V_k$ is the complex bus voltage, $Y$ the admittance matrix, and $\odot$ element-wise multiplication. Statistical significance is assessed using paired Wilcoxon signed-rank tests on $N_{\text{wilcox}}$ stratified samples with Bonferroni correction (Table~\ref{tab:main_results}) \cite{wilcoxon1945individual,demsar2006statistical}. We report both significance markers and effect sizes (e.g., $\Delta$RMSE, $\Delta\mathcal{L}_{\mathrm{PF}}$).

\subsection{Cross-Regime Generalization Performance (MV$\to$HV)}
\label{subsec:headline}
Zero-shot transfer yields $\mathrm{RMSE}_{\text{all}} = 1.65\times 10^{-2}$, indicating severe MV$\to$HV domain shift, consistent with prior findings on regime sensitivity of learned power-flow surrogates \cite{donon2020neural,lin2024powerflownet}. Full fine-tuning reduces error to $9.35\times 10^{-4}$, establishing the adaptation upper bound. 

\texttt{LoRA+PHead} achieves $\mathrm{RMSE}_{\text{all}} = 1.20\times 10^{-3}$, within $2.6\times 10^{-4}$ of Full FT while reducing trainable parameters by $85.46\%$ (Table~\ref{tab:main_results}), supporting \textbf{contribution~(i)} \cite{hu2022lora}. \texttt{Head-only} under-adapts ($1.38\times 10^{-3}$), while \texttt{LoRA-only} degrades further ($3.61\times 10^{-3}$), indicating that attention-only low-rank updates cannot fully compensate regime-dependent rescaling, consistent with the role of attention projections in GAT message passing \cite{velickovic2018graph}.

\subsection{Physical Consistency and Source-Domain Retention}
\label{subsec:physics_stability_retention}

Full FT achieves $\mathcal{L}_{\mathrm{PF}}=1.11$, while \texttt{LoRA+PHead} attains $1.21$ ($+9\%$, $\Delta\mathcal{L}_{\mathrm{PF}}=0.10$), demonstrating that parameter-efficient adaptation preserves physical feasibility at convergence, consistent with \textbf{contribution~(ii)}. In contrast, \texttt{LoRA-only} incurs a large residual ($\mathcal{L}_{\mathrm{PF}}=6.08$), indicating physically implausible solutions under insufficient adaptation capacity, in line with prior work on physics-informed regularization \cite{hu2021physics,dejongh2022physics,jeddi2021physics}. Training trajectories of $\mathcal{L}_{\mathrm{PF}}$ (Fig.~\ref{fig:physics_loss_full_ft}, Fig.~\ref{fig:physics_loss_lora_head}) show stable convergence for both \texttt{Full FT} and \texttt{LoRA+PHead} without divergence under heavily loaded regimes near voltage stability limits, indicating that restricting updates to low-rank subspaces does not degrade optimization stability for stiff AC power-flow losses \cite{bergen1999power}.

Stability--plasticity is quantified by the source-retention score $\mathcal{R}_{\text{ret}}$, defined as source-domain RMSE relative to the zero-shot MV baseline (100\% indicates no forgetting). \texttt{LoRA+PHead} balances strong target-domain adaptation with substantial source retention, whereas \texttt{Head-only} retains more but under-adapts, validating \textbf{contribution~(iii)} \cite{mccloskey1989catastrophic}. Unlike feature-space DA (e.g., CORAL \cite{sun2016deep}) that aligns intermediate distributions, our scheme operates in weight space on a low-rank submanifold around the source backbone --- an orthogonal axis composable with feature-space methods. Transfer to real operational SCADA data via \texttt{OPFData} \cite{lovett2024opfdata} or \texttt{PF}$\Delta$ \cite{bhagavathula2025pf} is the natural next step.

\subsection{Few-Shot Adaptation and Sample Efficiency}
\label{subsec:fewshot}

We analyze performance as a function of labeled target fraction $\beta$ (Fig.~\ref{fig:fewshot_band}). In the extreme low-shot regime ($\beta \leq 5\%$), low-rank adaptation lags Full FT due to subspace-constrained bias under limited supervision \cite{geman1992biasvariance}. For $\beta \geq 10\%$, \texttt{LoRA+PHead} approaches Full FT accuracy, characterizing the regime where parameter-efficient adaptation becomes Pareto-competitive (Fig.~\ref{fig:pareto_tradeoff}) and supporting \textbf{contribution~(iii)}.

\subsection{Efficiency--Accuracy Trade-off}
\label{subsec:pareto}

The Pareto analysis (Fig.~\ref{fig:pareto_tradeoff}) relates RMSE to trainable-parameter fraction $\rho$. Full FT ($\rho=1$) minimizes error at maximal cost; \texttt{LoRA-only} ($\rho\approx 3.4\%$) underfits. \texttt{LoRA+PHead} ($\rho\approx 14.5\%$) lies on the Pareto frontier, achieving near-Full-FT accuracy with a $6$--$7\times$ reduction in trainable parameters, reinforcing \textbf{contribution~(i)} and \textbf{contribution~(iii)}.

\section{Conclusion}
\label{sec:conclusion}

We addressed MV$\to$HV AC power-flow prediction via parameter-efficient adaptation of physics-informed self-attention GNNs. Restricting updates to low-rank perturbations of attention projections with selective head unfreezing, \texttt{LoRA+PHead} recovers near--full fine-tuning accuracy ($\Delta$RMSE $=2.6\times10^{-4}$) with $>85\%$ fewer trainable parameters and comparable physics residuals, at a 4.7~p.p.\ source-retention gap. These results demonstrate a controllable stability--plasticity trade-off for scalable deployment of learning-based AC-PF solvers.

\renewcommand{\baselinestretch}{0.94}\selectfont
\bibliographystyle{IEEEtran} 
\bibliography{refs}          
\end{document}